\documentclass[10pt]{article}
\usepackage[a4paper, margin=1in]{geometry}
\usepackage[T1]{fontenc}
\usepackage[normalem]{ulem}

\usepackage[dvipsnames,svgnames,x11names]{xcolor}

\colorlet{RED}{red}

\colorlet{WHITE}{white}
\colorlet{FORESTGREEN}{ForestGreen}
\colorlet{MAROON}{Maroon}

\newcommand{\agen}{AeroGen\xspace}
\newcommand{\CLG}{CLG\xspace}

\newcommand{\aeval}{AeroEval\xspace}

\usepackage[nocompress]{cite}
\usepackage{balance}

\usepackage{listings}

\lstdefinelanguage{XML}
{
basicstyle=\ttfamily\footnotesize,
  morestring=[b]",
  moredelim=[s][\bfseries\color{Maroon}]{<}{\ },
  moredelim=[s][\bfseries\color{Maroon}]{</}{>},
  moredelim=[l][\bfseries\color{Maroon}]{/>},
  moredelim=[l][\bfseries\color{Maroon}]{>},
  morecomment=[s]{<?}{?>},
  morecomment=[s]{<!--}{-->},
  commentstyle=\color{gray},
  stringstyle=\color{blue},
  identifierstyle=\color{red}
}

\usepackage{moreverb}

\usepackage[nounderscore]{syntax}

\usepackage[pdftex]{graphicx}
\graphicspath{{./figures/}}
\DeclareGraphicsExtensions{.pdf}

\usepackage[cmex10]{amsmath}
\usepackage{amssymb}
\usepackage{mathtools}
\usepackage{amsthm}
\usepackage{amsfonts}
\usepackage{gensymb}

\usepackage[caption=false,font=footnotesize]{subfig}
\usepackage{algorithmicx}
\usepackage{algpseudocode}
\usepackage[ruled]{algorithm}
\definecolor{light-gray}{gray}{0.75}
\algrenewcommand{\algorithmiccomment}[1]{\hskip3em{{\footnotesize \textcolor{light-gray}{$\blacktriangleright$}}} #1}

\usepackage{subcaption}

\usepackage{multirow}
\usepackage{rotating} 
\usepackage{booktabs} 
\usepackage{colortbl} 
\usepackage{tablefootnote}

\usepackage{array}
\newcolumntype{L}[1]{>{\raggedright\let\newline\\\arraybackslash\hspace{0pt}}m{#1}}
\newcolumntype{C}[1]{>{\centering\let\newline\\\arraybackslash\hspace{0pt}}m{#1}}
\newcolumntype{R}[1]{>{\raggedleft\let\newline\\\arraybackslash\hspace{0pt}}m{#1}}

\usepackage{xurl}

\usepackage[pdftex,colorlinks=true,urlcolor=blue,citecolor=blue]{hyperref}

\usepackage{orcidlink}

\usepackage{xspace}

\usepackage{enumitem}

\usepackage{blindtext}

\begin{document}

\title{
\aeval: Staged Program and Execution Validation for AI-Generated Drone Missions
}

\author{
\normalsize
Kautuk Astu~\orcidlink{0009-0000-6268-811X},
Naina Rabha~\orcidlink{0009-0005-3089-3879} and
Yogesh Simmhan~\orcidlink{0000-0003-4140-7774}\\
\normalsize\textit{Department of Computational and Data Sciences,}\\
\normalsize\textit{Indian Institute of Science, Bangalore 560012, India}\\
\normalsize\texttt{\{kautukastu, simmhan\}@iisc.ac.in}
}

\date{}

\maketitle

\pagestyle{plain}

\begin{abstract}
Large Language Models (LLMs) can generate drone programs from natural-language mission descriptions, but syntactically valid programs may still violate user intent, environmental constraints, and mission-level behavior. This problem is pronounced in cyber-physical applications, where correctness depends on the interaction among generated code, mobile sensing, environmental geometry, event-driven analytics, and physical execution. Existing drone code-generation systems primarily use prompt guardrails or simulator outcomes and provide limited failure localization. We present \textit{\aeval}, an agent-assisted middleware for staged validation of AI-generated drone missions. \aeval combines deterministic program analysis with context-grounded LLM agents. It first validates program syntax, platform API usage, and mission intent, and then evaluates the realized behavior using execution trajectories, mission requirements, and environmental context. Each stage returns structured failure information for iterative regeneration. In our evaluation using 20 navigation tasks and five analytical mission types over AirSim and Gazebo simulators, \aeval improves navigation success from $55\%$ to $95\%$. In a stagewise ablation study, our Code and Trajectory Validators by themselves achieve mean run-level success rates of $44\%$ and $56\%$, respectively, while the full \aeval pipeline achieves $88\%$; the stages detect complementary failures in program structure, API usage, mission intent, obstacle avoidance, altitude, coverage, and event-driven transitions and the guided regeneration corrects for them. Across the main analytics missions, \aeval increases aggregate run-level success from $34\%$ for one-shot \agen to $88\%$ within the regeneration budget. These results demonstrate the benefit of combining program-level and execution-grounded agentic validation for AI-generated drone applications in the evaluated environment.
\end{abstract}

\section{Introduction}
\label{sec:intro}
Large Language Models (LLMs) allow users to construct drone applications by specifying missions such as delivery, survey, inspection, and target tracking in natural language~\cite{liang-icra-2023,gsce,wang-icra-2026,aerogen}. The generated programs coordinate navigation, mobile sensing, analytics, and control across drone, edge, and cloud resources. Their correctness therefore depends on more than syntactic validity. A program must use the available platform interfaces, preserve user intent, respond to sensed events, and satisfy spatio-temporal constraints imposed by the physical environment. AI-generated drone missions are consequently pervasive cyber-physical applications whose software behavior must be interpreted relative to their execution context~\cite{10.1145/3486607.3486750,9612021,uav-crowd-percom24, dronevlc-percom23}.

However, existing code generation workflows provide limited evidence that a \textit{plausible program} actually realizes its requested mission. A syntactically valid program may fly directly through an obstacle, survey a region along the wrong orientation, leave part of a search region uncovered, or fail to transition from search to tracking after detecting a target. These errors depend on environmental geometry, sensing events, and realized trajectories, and may \textit{remain undetected} by code-level checks. Conversely, a simulator failure alone \textit{does not identify} whether the cause is an invalid API call, an intent mismatch, or incorrect physical behavior. Hence, validating generated drone applications requires \textit{both program-level and execution-grounded evidence}.

Existing \textit{cyber-physical code-generation systems} provide only part of this lifecycle. \textit{Structured prompting} improves generated drone code but discovers failures primarily through subsequent execution~\cite{gsce}. \textit{Simulation-guided generation} has also been applied to autonomous-driving software, where rule-based assessments of critical traffic scenarios are returned to an LLM for code repair~\cite{nouri-simulation-guided-2025}. \CLG similarly uses \textit{simulator feedback} to iteratively refine drone navigation programs~\cite{wang-icra-2026}, while \agen \textit{combines planning, code generation, sensing, and analytics} for more complex missions~\cite{aerogen}. 
However, these systems do not \textit{systematically separate failures} caused by program structure, platform API usage, mission interpretation, and realized physical behavior. \textit{Validation frameworks} for autonomous systems combine formal specifications, simulation, digital twins, and domain-specific tests~\cite{araujo-ras-vnv-2023,polyverif-2023}, but typically assume that the application logic and testable properties have already been authored. In contrast, an \textit{AI-generated mission} is an untrusted program synthesized from an informal user specification~\cite{safety-llm-driven-robotics,robotics-sw-engg}.

We present \textbf{\aeval}, an \textit{agent-assisted middleware} that performs staged validation between AI-based mission generation and drone execution. Its \textit{Code Validator} combines deterministic syntax and API checks with context-grounded validation of mission intent. Programs that pass this stage execute in a \textit{physics-based simulator}. The \textit{Trajectory Validator} then evaluates the realized behavior using the mission specification, robot and world descriptions, execution trajectory, and observed events. For \textit{navigation tasks} with a reference path, \aeval uses numerical trajectory comparison. For \textit{analytical missions} without a unique reference path, a bounded validation agent evaluates properties such as obstacle avoidance, altitude, coverage, orientation, and event-driven transitions. A failed stage returns a \textit{structured diagnosis to the generator for refinement}.

\begin{figure}[!t]
\centering
    \subfloat[Run-level success for \agen and \aeval
    \label{fig:run-level-success}]{
    \includegraphics[width=0.35\columnwidth]{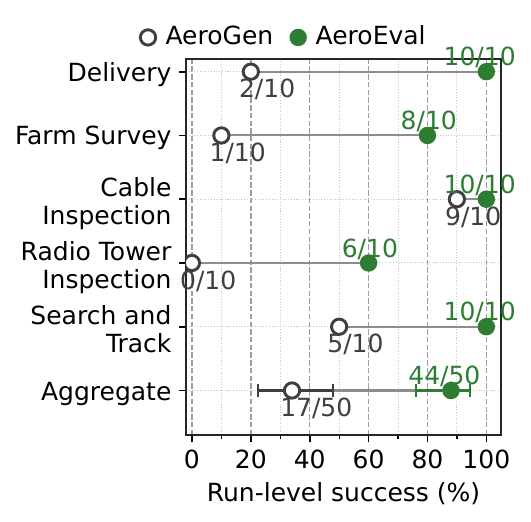}
    }
    \qquad
    \subfloat[Mean run-level success of \aeval configurations.
    \label{fig:stagewise-composition}]{%
        \includegraphics[width=0.35\columnwidth]{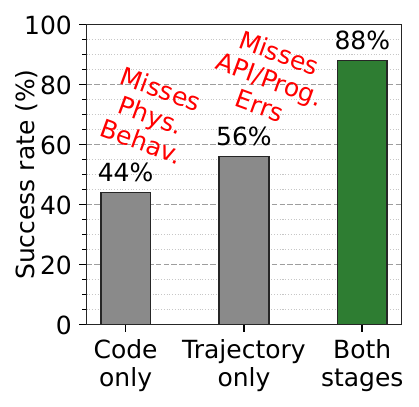}
    }%
\caption{Impact and composition of staged mission validation. (a) Run-level success for one-shot \agen and \agen+\aeval over 10 independent runs per analytical mission using o3-mini; aggregate error bars show Wilson score $95\%$ confidence intervals. (b) Mean run-level success of the Code Validator-only, Trajectory Validator-only, and full \aeval pipeline with a budget of up to three regenerations.}
\label{fig:motivation}
\end{figure}

The key novelty of our work is the \textit{composition} of program-level and execution-grounded validation within a common mission lifecycle. Figure~\ref{fig:motivation} summarizes both its impact and the reason for staged validation. Under a common budget of at most three regeneration cycles, the Code and Trajectory Validators alone achieve mean run-level success rates of $44\%$ and $56\%$, respectively, while their composition in the complete \aeval workflow achieves $88\%$. Relative to the underlying generators, \aeval improves navigation success over the closed-loop \CLG baseline from $11/20$ to $19/20$ tasks, and analytical run-level success over one-shot \agen from $17/50$ to $44/50$ runs. This is because program validation detects structural, API, and intent errors, while execution-grounded validation detects violations that arise from the interaction between the generated program and its physical environment.

In this paper, we make the following key contributions: 
\begin{enumerate}[leftmargin=*] 
\item We propose a \textit{staged validation lifecycle} that separates program validity, mission-intent alignment, and realized physical behavior, and returns stage-specific feedback for iterative regeneration (Sec.~\ref{sec:design}).

\item We develop \textit{context-grounded behavioral validation} for analytical missions without a unique reference trajectory. The \textit{validation agents} combine mission requirements, program artifacts, robot and world descriptions, and execution observations to evaluate spatio-temporal and event-driven properties (Sec.~\ref{subsec:trajectory-validation}).

\item We evaluate \aeval across \textit{20 navigation tasks and five analytical mission types} using two generators, two simulators and two reasoning models. The evaluation includes stagewise ablations, failure localization, limited physical-drone execution, and validation overheads (Sec.~\ref{sec:eval}). 
\end{enumerate}

\section{Related Work}
\label{sec:related-work}
\aeval lies at the intersection of AI-based drone program synthesis, validation of autonomous systems, and middleware for pervasive cyber-physical applications. Prior work has studied these concerns largely independently. We instead examine how program analysis, mission context, and execution observations can be combined to validate and repair AI-generated drone applications.

\subsubsection{AI-based drone program synthesis} 
LLMs can translate natural-language specifications into executable programs for embodied control~\cite{liang-icra-2023}. In the drone domain, GSCE uses structured prompts, API constraints, and examples, but detects failures primarily through subsequent execution~\cite{gsce}. \CLG incorporates simulator observations into a refinement loop for navigation tasks, regenerating programs that violate waypoint, altitude, yaw, or geometric requirements~\cite{clgsce,wang-icra-2026}. \agen extends single-shot generation to analytical missions that compose navigation, sensing, analytics, and deployment over heterogeneous resources~\cite{aerogen}.

Simulation-guided generation has also been applied to autonomous-driving software, where generated control code is evaluated over predefined critical scenarios and rule-based assessments are returned for repair~\cite{nouri-simulation-guided-2025}. Corrective robot-code generation similarly uses physics-based execution or static text-based simulation to predict navigation failures and refine generated operations~\cite{wang-icra-2026}. These approaches demonstrate the value of execution feedback, but organize validation around scenario outcomes or deviations from specified navigation behavior.

\aeval instead proposes a validation lifecycle that separates deterministic syntax and API checks, context-grounded intent validation, and post-execution behavioral validation. This distinction is necessary because invalid API usage may not be observable from a trajectory, while collision, coverage, orientation, and event-transition violations may not be inferred reliably from program text. Further, \aeval supports analytical missions with multiple valid trajectories by evaluating mission-specific spatio-temporal and event-driven properties using the mission specification, generated program, environment descriptions, and execution trace.

\subsubsection{Validation of autonomous systems} Validation and Verification of autonomous systems commonly use formal specifications, simulation, digital twins, scenario-based testing, and runtime monitoring~\cite{araujo-ras-vnv-2023, polyverif-2023,torens2013automated,matalonga2024verification, aslandere2025survey}. PolyVerif treats validation of AI-enabled Cyber-Physical Systems (CPS) as an infrastructure problem requiring reusable simulators, test environments, and analysis tools~\cite{polyverif-2023}. Co-zyBench similarly uses co-simulation middleware and Digital Twins to evaluate context-dependent behavior before physical deployment~\cite{cozybench-percom24}. 
These techniques typically assume that application logic has been authored and that the evaluated properties can be formally or procedurally specified. In contrast, an AI-generated mission is an untrusted program synthesized from an informal user specification and may misinterpret the task before controller-level properties can be examined.
\aeval instead validates program structure, platform interfaces, mission intent, and realized spatio-temporal behavior before physical execution; formal verification and runtime assurance remain complementary.

LLM-guided validation has also been used to translate natural-language driving policies into checkable specifications, analyze their consistency with autonomous-driving software, and generate simulation tests~\cite{zhang-semantic-validation-2026}.
This approach validates software against explicit safety policies,
whereas \aeval localizes failures across generated program structure,
mission intent, and realized behavior for navigation and analytical
drone missions.

\subsubsection{Middleware for pervasive drone applications} Middleware and runtime systems support mobile applications across robots, sensors, edge devices, and cloud resources. Service-oriented Drones-as-a-Service (DaaS) systems provide abstractions for navigation, sensing, analytics, scheduling, and data management~\cite{aerodaas}, while microservice-based platforms coordinate mission planning and resource allocation across multiple drones~\cite{percom-daas-microservice,percom-multi-drone-control}. Context-adaptive middleware such as Furcifer selects among local, edge, split, and cloud execution according to network and system conditions~\cite{furcifer-percom24}, while OctopInf performs workload-aware scheduling for edge video-analytics pipelines~\cite{octopinf-percom25}. Aerial-edge systems similarly combine UAV mobility with communication and computation placement~\cite{dt-aerial-edge-tvt22}. These systems optimize the execution of supplied applications, whereas \aeval examines whether AI-generated mission logic satisfies its program-level and physical-context requirements before execution.

Drones also operate as mobile sensing and analytics platforms in
context-dependent environments. Prior systems perform on-board
crowd-motion analysis under changing viewpoints~\cite{uav-crowd-percom24} and use drones to collect data from
batteryless sensors under environment-dependent communication
conditions~\cite{dronevlc-percom23}. These applications illustrate that
drone behavior depends jointly on mobility, sensing, communication,
computing resources, and physical context, but assume that application
logic is authored and suitable for execution.

\subsubsection{Summary} Prior work has developed AI-based drone generators, validation techniques for autonomous systems, and middleware for context-aware mobile applications. However, generators provide limited failure localization, validation techniques typically assume authored application logic, and execution middleware assumes that supplied programs are suitable for execution. \aeval connects these areas through a staged lifecycle that separates program validity, mission-intent alignment, and realized physical behavior. Its technical distinction is context-grounded validation of analytical missions without a unique reference trajectory, together with stage-specific feedback for iterative repair.

%%%%%%%%%%%%%%%%%%%%%%%%%%%%%%%%%%%%%%%%%%%%%%%%%%%%%%%%%%%%%%%%%%%%%%%%
%%%%%%%%%%%%%%%%%%%%%%%%%%%%%%%%%%%%%%%%%%%%%%%%%%%%%%%%%%%%%%%%%%%%%%%%
\section{\aeval Design}
\label{sec:design}

\begin{figure}[t]
\centering
\includegraphics[width=0.65\columnwidth]{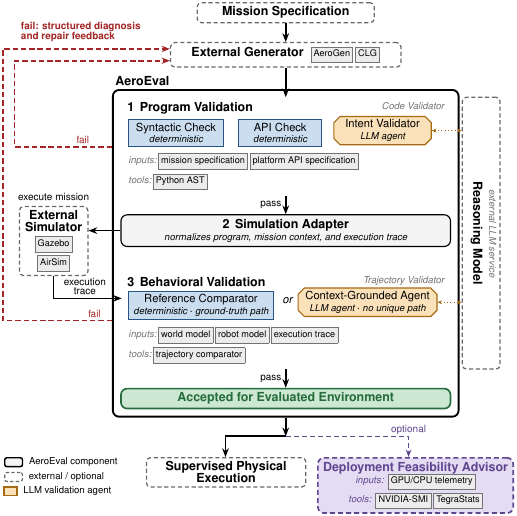}
\caption{\aeval architecture and validation lifecycle. Program
Validation combines deterministic syntax and API checks with an
agent-based intent check. Programs that pass execute in an external
simulator, and the resulting trace is evaluated using either a
reference-path comparator or a context-grounded validation agent.
Failures return stage-specific diagnoses to the generator. Acceptance
is scoped to the evaluated mission and environment.
Dashed components and links are external or optional.}
\label{fig:architecture}
\end{figure}

\subsection{Architecture and Lifecycle} \label{subsec:architecture} Figure~\ref{fig:architecture} shows \aeval operating between an AI-based mission generator and the drone execution environment. Its input is a candidate \textit{bundle} containing the natural-language mission specification, generated program, platform interfaces, and robot and world descriptions. The middleware processes this bundle through Program Validation, simulated execution, and Behavioral Validation before accepting the candidate for the evaluated environment.

\textit{Program Validation} checks syntax, platform API usage, and alignment with mission intent. A candidate that passes executes in a physics-based simulator, whose adapter returns a normalized execution record containing the trajectory, vehicle state, mission events, and environment interactions. \textit{Behavioral Validation} evaluates this record against the spatial, temporal, and event-driven properties derived from the mission specification. Thus, each stage operates over distinct evidence: program artifacts before execution and observed mission behavior after execution.

Each validator returns a record consisting of $\langle$stage, outcome, violation, evidence, correction$\rangle$. A failed record is supplied to the generator together with the original candidate bundle. The regenerated program re-enters Program Validation because a behavioral correction may introduce a syntax, API, or intent error. The lifecycle terminates when a candidate \textit{passes} both mandatory stages or exhausts the configured \textit{refinement} budget, in which case \aeval exits the feedback loop and refers the run for human intervention.

A candidate accepted by \aeval has passed the implemented program, intent, simulator-execution, and behavioral checks for the evaluated mission and environment. 
This acceptance provides \textit{empirical validation evidence}; it \textit{does not provide a formal guarantee} of safety, liveness, adversarial robustness, or correctness under untested conditions. Physical execution remains supervised and requires the robot, platform, and environmental assumptions used during validation to remain applicable.

\subsection{Agents, Context, and Tools} \label{subsec:tools}

\aeval combines deterministic analysis with bounded validation agents. We use the term \textit{validation agent} for an LLM-based component that performs a specific validation task using structured instructions, mission context, and tool or execution observations. The agent does not autonomously modify or execute mission code. Instead, it returns a \textit{structured validation record} to the upstream generator, which performs the regeneration.

The \textit{Intent Validator} receives the mission specification, generated program, permitted platform APIs, and intent-validation criteria. The \textit{Context-Grounded Agent} additionally receives the robot and world descriptions, executed trajectory, mission events, and behavioral criteria. The robot description captures platform constraints such as maximum flight speed and altitude, payload capacity, and battery limitations, while the world description specifies environmental constraints including obstacles, the road network, permitted flying corridors, no-fly regions, and maximum allowed height. These inputs ground validation in the user request, platform capabilities, physical environment, and observed behavior, rather than requesting an unconstrained judgment from the reasoning model.
Table~\ref{tab:validation-properties} summarizes the evidence and mechanism used to evaluate each class of mission property.

\textit{Deterministic tools} complement the validation agents. The current implementation uses the Python Abstract Syntax Tree (AST) library for parsing and structural inspection~\cite{python-ast}, platform API metadata for interface checks, and a numerical comparator for navigation tasks with a reference trajectory~\cite{clgsce}. Gazebo and AirSim provide physics-based execution~\cite{gazebo,airsim2017fsr} and expose trajectories, vehicle state, and mission events through different interfaces. \aeval adapters normalize these observations before passing them to the Trajectory Validator.

Our evaluations use o3-mini and Gemini-3.1-Pro as reasoning models for our agents~\cite{gpt-o3-mini, gemini-3.1-pro}. The default configuration uses o3-mini for both generation and validation, with distinct task instructions and evidence for each role. Supporting another programming language just requires a parser and platform-specific API metadata, but does not change the interfaces used by the validation agents.

\subsection{Code Validator} \label{subsec:code-validation}

The \textit{Code Validator} determines whether a generated program is structurally executable, uses the available platform interfaces, and represents the requested mission before simulator resources are allocated. It contains syntactic, API, and intent checks. The first two use deterministic analysis where possible, while the Intent Validator uses a context-grounded validation agent.

The \textit{Syntactic Validator} parses the generated program and 
identifies malformed syntax and invalid constructs that prevent execution.
For the evaluated Python drone programs, it uses the AST library and lightweight structural inspection~\cite{python-ast}. A failed check returns the parser diagnostic and source location to the generator. Syntax parsing is prior capability leveraged by \aeval; its role is to reject invalid programs before invoking simulation or additional validation agents.

The \textit{API Validator} examines calls to the drone, simulator, sensing, and analytics interfaces available on the target platform. 
It deterministically checks operation names and dependencies on unsupported services using program structure and platform API metadata, while an agent evaluates call ordering relative to the surrounding mission logic.
These checks detect generated calls that appear plausible but refer to nonexistent operations or unsupported interfaces.

The \textit{Intent Validator} evaluates whether the generated program represents the requirements in the natural-language mission specification. It extracts the required mission phases, constraints, and event-driven transitions, and examines whether the program contains corresponding logic. For example, it can identify a delivery program that omits a destination or the return-to-depot phase. However, it does not determine from program text whether the realized trajectory avoids every obstacle or provides the required spatial coverage. Such properties are evaluated after simulation (Sec.~\ref{subsec:trajectory-validation}).

A candidate advances to simulated execution only after passing all three
checks. Otherwise, \aeval combines the detected violations and
corrective feedback into a validation record and returns the candidate
to the generator. Passing the Code Validator indicates program-level
validity under the implemented checks; it does not indicate that the
program realizes the required physical behavior.

\begin{table}[t] \centering \caption{Evidence and validation mechanisms used for program-level and
execution-grounded mission properties.} \label{tab:validation-properties} \small 
\begin{tabular}{@{}lll@{}} \toprule \bf Property & \bf Evidence & \bf Validator \\ \midrule Syntax & Program & Deterministic \\ API availability & Program, API metadata & Deterministic \\ Mission phases & Program, specification & Intent Agent \\ Reference path & Waypoints & Comparator \\ Obstacle/corridor & Trace, world context & Context Agent \\ Coverage/orientation & Trace, mission region & Context Agent \\ Event transition & Events, trace & Context Agent \\ \bottomrule \end{tabular} \end{table}

\subsection{Trajectory Validator}
\label{subsec:trajectory-validation}

The \textit{Trajectory Validator} evaluates whether a program that
passes code validation realizes the required behavior in the simulated
environment.
\aeval records the executed trajectory, vehicle state, mission events, and environment interactions, and then applies one of two validation modes depending on whether the
evaluation provides a reference trajectory.

For navigation tasks, \aeval uses a \textit{numerical Trajectory Comparator} to evaluate the observed path against a reference trajectory~\cite{wang-icra-2026}.
The evaluated CLG tasks assume deterministic waypoint sequences, therefore, the observed trajectory must contain the same number of waypoints as the reference trajectory, and any mismatch is treated as a failure. For matching-length trajectories, a $2$\,meter (configurable) positional tolerance is applied independently to each corresponding waypoint, together with mission-specific terminal conditions.
This comparator is adopted from the \CLG workflow.

Analytical missions may admit multiple valid trajectories because the path can depend on the environment, sensing outcomes, and event locations that feed through DNNs or other ML models for online decisions. For these missions, \aeval uses a \textit{Context-Grounded Agent} within the Trajectory Validator. 
The agent receives the mission specification, generated program, executed trajectory, robot and world descriptions, observed mission events, and behavioral criteria. 
The executed trace is represented as an ordered list of odometry tuples $(x,y,z,\mathrm{yaw},\mathrm{hover\_duration})$.
Obstacles and mission regions are described in natural language within the world-information guardrail. Using these inputs, the Trajectory Validation Agent evaluates properties such as obstacle and corridor compliance, altitude bounds, spatial coverage, camera orientation, waypoint completion, and event-driven transitions. 
The current implementation does not use separate geometric helpers; the agent infers these properties directly from the textual world
description and execution trace.

This validation differs from reference-path comparison because it determines whether the observed \textit{behavior satisfies mission properties} rather than whether the drone reproduces one expected path. For example, a survey may use different valid waypoint sequences while still covering the required region with the specified orientation and lane spacing. Similarly, the location of a transition from search to tracking may depend on when the target is detected.

Each behavioral criterion produces a \textit{pass} or \textit{fail} outcome, supporting execution observations, and corrective feedback. A failed candidate and its validation record return to the generator, while the regenerated candidate re-enters the lifecycle at the Code Validator. This prevents a behavioral correction from introducing an undetected syntax, API, or intent failure. 

The agent output provides empirical validation, and later we evaluate its false acceptances, false rejections, repeated-trace consistency, and agreement across reasoning models in Sec.~\ref{subsec:ablation}.

\subsection{Optional Deployment Advisor} \label{subsec:optimizer}

After a candidate passes the mandatory validation stages, an optional advisor can use resource telemetry to recommend alternative analytics placements or waypoint schedules. For example, the advisor may identify that concurrent analytical models exceed the memory available on an edge device, or that waypoint ordering can be shortened without changing mission requirements. These recommendations do not determine mission acceptance and are returned to the generator or runtime scheduler rather than applied directly.

The present evaluation considers this component only as an extensibility
mechanism under constructed resource contention. The principal
contribution and evaluation of \aeval concern program-level and
execution-grounded mission validation.

\section{Evaluation}
\label{sec:eval}

We evaluate \aeval along four dimensions: (i) whether its validation
stages identify complementary failure classes, (ii) whether the
agent-based validators provide consistent decisions, (iii) whether the
complete lifecycle improves mission success, and (iv) whether its
latency and model costs remain practical. We use navigation missions
with reference trajectories and analytical missions whose valid
behavior depends on sensing, environmental context, and event-driven transitions.

\subsection{Methodology}
\label{subsec:methodology}

\begin{table}[t]
\centering
\caption{Summary of the experimental configurations.}
\label{tab:experimental-setup}
\small
\setlength{\tabcolsep}{5pt}
\begin{tabular}{@{}lllll@{}}
\toprule
\bf Study & \bf Generator & \bf Environment & \bf Missions & \bf Budget \\
\midrule
Navigation
    & \CLG
    & AirSim
    & 20 tasks
    & 15 iterations \\
Analytical
    & \agen
    & Gazebo
    & 5 types $\times$ 10 runs
    & 15 iterations \\
Ablation
    & \agen
    & Gazebo
    & 5 types $\times$ 10 runs
    & 3 regenerations \\
Physical
    & \agen
    & Tello
    & 3 types
    & supervised \\
\bottomrule
\end{tabular}
\end{table}

\subsubsection{Workloads}
We evaluate two mission families with different behavioral
requirements. The \textit{navigation family} contains 20 tasks from \CLG,
including waypoint traversal, altitude control, yaw control, and
geometric flight patterns~\cite{wang-icra-2026}. These tasks have reference
trajectories and are executed using AirSim. The \textit{analytical family}
contains five mission types from \agen~\cite{aerogen}: multi-destination delivery, farm survey, cable inspection, radio-tower inspection, and search and track~\cite{package-delivery,farm-survey,cable-inspection,radio-inspection, target-tracking}.
These missions combine navigation with sensing,
analytics, spatial coverage, and event-driven behavior, and are executed
using Gazebo.

Each analytical mission type is evaluated using 10 \textit{independent runs}.
The main reliability experiment permits at most 15 refinement iterations
per run. A run that does not produce an \textit{accepted candidate} within this
budget terminates and is referred for human intervention. To control the
cost of the stagewise comparison, all ablation configurations instead
use one initial candidate and at most three regeneration cycles per run.

\subsubsection{Experimental environment}
The analytical experiments execute on a workstation with an NVIDIA
RTX~3090 GPU with 24\,GB VRAM and 124\,GB RAM. We use Gazebo Ignition
Fortress~\cite{gazebo} with a PX4-based quadrotor in a custom environment containing
roads, buildings, farms, and radio towers. Analytical services include
YOLOv8-based object detection~\cite{10533619}, ORB-based tower detection, and
PID-based tracking, and execute in isolated containers. The navigation
experiments use \CLG and AirSim~\cite{airsim2017fsr} on the same workstation.

The default reasoning model is o3-mini for both generation and
agent-based validation. Generation and validation use distinct
instructions and evidence, but may retain correlated model errors.
We additionally evaluate Gemini-3.1-Pro to study sensitivity to the
reasoning model.
The prompts and reasoning configuration remain fixed across independent
runs. The reasoning models do not expose temperature or conventional
sampling controls for the evaluated configuration, and we use the
default medium reasoning setting. One obstacle parameter in the world
description varies across runs and is appended near the end of the input
context; the remaining prompt and mission guardrails are unchanged.

\subsubsection{Metrics}
We distinguish pipeline convergence from the quality of individual
generated candidates. The \textit{run-level success rate},
$SR_{\mathrm{run}}$, is the fraction of independent runs that produce a
candidate accepted by both mandatory validation stages within the
configured refinement budget:
\begin{equation}
SR_{\mathrm{run}} =
\frac{N_{\mathrm{converged}}}{N_{\mathrm{runs}}}.
\label{eq:sr-run}
\end{equation}
The \textit{candidate-level success rate},
$SR_{\mathrm{candidate}}$, is the fraction of individual generated
programs accepted by both stages:
\begin{equation}
SR_{\mathrm{candidate}} =
\frac{N_{\mathrm{accepted\ candidates}}}
     {N_{\mathrm{generated\ candidates}}}.
\label{eq:sr-candidate}
\end{equation}
The former measures whether the iterative lifecycle eventually succeeds,
while the latter characterizes the quality of individual generated
programs. For converged runs, we also report iterations-to-success,
end-to-end latency, and token usage.
We define an iteration as one code-generation event within an independent run. The initial candidate is counted as iteration $1$. Thus, a budget of three regenerations permits up to four iterations per run. We report 95\% confidence intervals for $SR_{\mathrm{run}}$ using the Wilson score interval for a binomial proportion.

The stagewise ablation evaluates the Code Validator-only,
Trajectory Validator-only, and complete \aeval configurations using
independent initial generations. Each configuration uses 10 independent
runs for each of the five analytical mission types, with a common budget
of at most three regeneration cycles per run. The experiment therefore
compares configuration-level outcomes over repeated runs rather than
paired outcomes for identical initial candidates.

\subsection{Stage Complementarity and Validator Dependability}
\label{subsec:ablation}

\begin{figure}[t]
    \centering
    \subfloat[Stagewise ablation.
    \label{fig:stage-ablation}]{%
        \includegraphics[width=0.4\columnwidth]
        {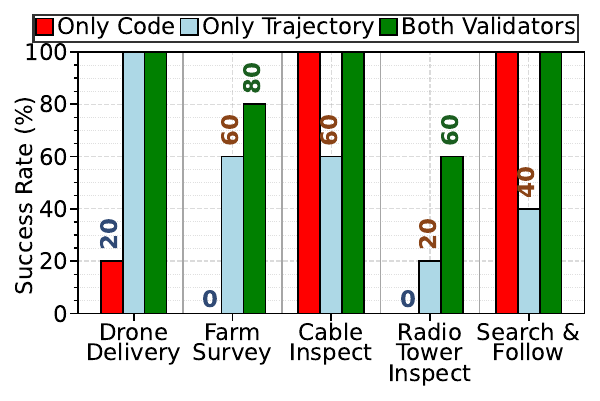}
    }%
    \qquad
    \subfloat[Validator dependability.
    \label{fig:validator-dependability}]{%
        \includegraphics[width=0.4\columnwidth]
        {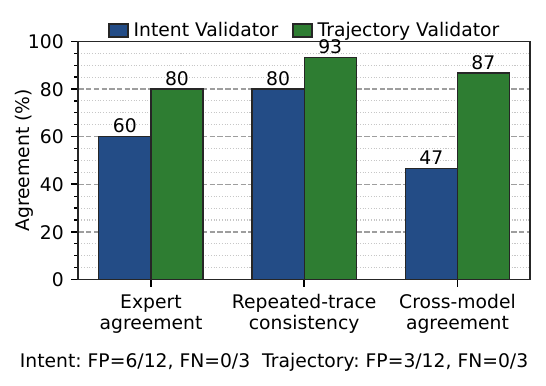}
    }%
    \caption{Contribution and dependability of the validation stages.
    Panel~(a) reports mean $SR_{\mathrm{run}}$ across five analytical mission types, with 10 independent runs per mission and a common limit of three regeneration cycles for each configuration. Panel~(b) reports expert agreement, repeated-trace consistency, and cross-model agreement for the Intent and Trajectory Validator; false-accept and false-reject counts are shown below the panel.}
    
    \label{fig:stage-validator-analysis}
\end{figure}

\subsubsection{Program-level and execution-grounded validation address complementary failure classes.} Figure~\ref{fig:stage-ablation} reports the stagewise ablation over five analytical mission types, with 10 independent runs per mission and a common budget of at most three regeneration cycles. The Code Validator-only and Trajectory Validator-only configurations achieve mean run-level success rates of $44\%$, ($95\%\ \text{CI}: 0.31 - 0.58$) and $56\%$, ($95\%\ \text{CI}: 0.42 - 0.69$), respectively, while the complete configuration achieves $88\%$ ($95\%\ \text{CI}: 0.76 - 0.94$). Thus, the complete configuration improves success by 44 percentage points over Code Validator-only and 32 percentage points over Trajectory Validator-only validation.

The three configurations use independently generated initial candidates rather than identical candidates shared across configurations. Consequently, the experiment compares configuration-level outcomes over repeated runs rather than paired outcomes for the same initial generation.

Without the Trajectory Validator, code-valid programs frequently violate physical mission requirements. Delivery programs take a direct path and collide with a building in $8/10$ runs, Farm Survey begins from the $+y$ edge rather than the required $+x$ edge in $10/10$ runs, and Radio Tower Inspection violates altitude or lane-spacing coverage requirements in $10/10$ runs. These failures survive program-level checks because they depend on the realized trajectory and its interaction with the environment.

Without the Code Validator, Farm Survey has one false acceptance by the Trajectory Validator and reaches the refinement limit in $3/10$ runs. Radio Tower Inspection uses insufficient altitude or excessive lane spacing in $8/10$ runs, resulting in incomplete coverage. Search and Track contains invalid API usage in $4/10$ runs and unnecessary yaw operations in $2/10$ runs. These failure categories are non-exclusive, and a single run may exhibit more than one failure mode. 

These observations explain the composition gain. Program validation
removes structural, interface, and intent failures before simulation,
while trajectory validation detects spatial and event-driven violations
that become observable only during execution.

\subsubsection{Many code-valid candidates subsequently fail behavioral validation.} Across the analytical missions, 12 Delivery, 27 Farm Survey, and 64 Radio Tower Inspection candidates pass the Code Validator but fail the Trajectory Validator. These outcomes are not false acceptances by the Code Validator because the two stages evaluate different properties. Instead, they quantify candidates whose syntax, API usage, and apparent intent alignment pass program-level validation, but whose realized behavior violates the mission requirements.

This distinction further explains the stagewise composition gain. Program validation rejects structural, interface, and intent failures before simulation, while behavioral validation detects violations that depend on trajectories, environmental geometry, spatial coverage, and mission events. Cable Inspection and Search and Track have no observed code-pass/behavior-fail candidates in the current results. 
The counts are computed per candidate and may include overlapping behavioral failure categories.

\subsubsection{The Trajectory Validator provides empirical rather than formal validation evidence.}

Against the current 15 expert-labelled outcomes, comprising 12 expert-rejected and 3 expert-accepted cases, it produces three false acceptances and no false rejections, corresponding to a false-accept rate of $3/12=25\%$ and a false-reject rate of $0/3=0\%$. Each trace is evaluated three times independently with o3-mini using the identical validator input; fourteen of the 15 traces receive the same pass/fail label in all three o3-mini repetitions, yielding $14/15=93\%$ all-run consistency.
We additionally evaluate each trace once with Gemini-3.1-Pro. The o3-mini and Gemini-3.1-Pro show $87\%$ raw cross-model agreement.
Given the small and imbalanced expert-labeled set, we report raw cross-model agreement rather than interpreting a chance-corrected coefficient.

The $25\%$ false-accept rate confirms that the Context-Grounded Agent cannot serve as a standalone assurance mechanism. \aeval instead combines execution-grounded agent decisions with deterministic checks, structured mission context, iterative repair, and human intervention for non-converged runs.

\subsection{End-to-End Mission Reliability}
\label{subsec:reliability}

\begin{figure}[t]
    \centering
    \subfloat[Navigation-task outcomes.
    \label{fig:navigation-reliability}]{
        \includegraphics[width=0.4\columnwidth]
        {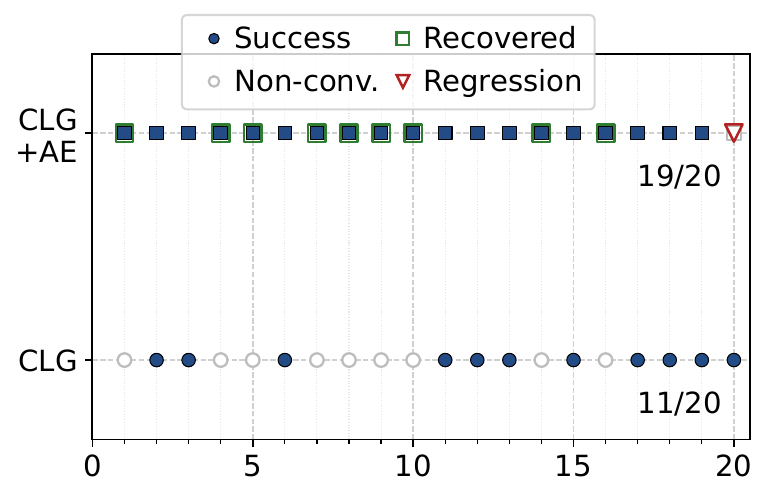}
    }
    \qquad
    \subfloat[Analytical-mission run-level success.
    \label{fig:analytical-run-success}]{
        \includegraphics[width=0.4\columnwidth]
        {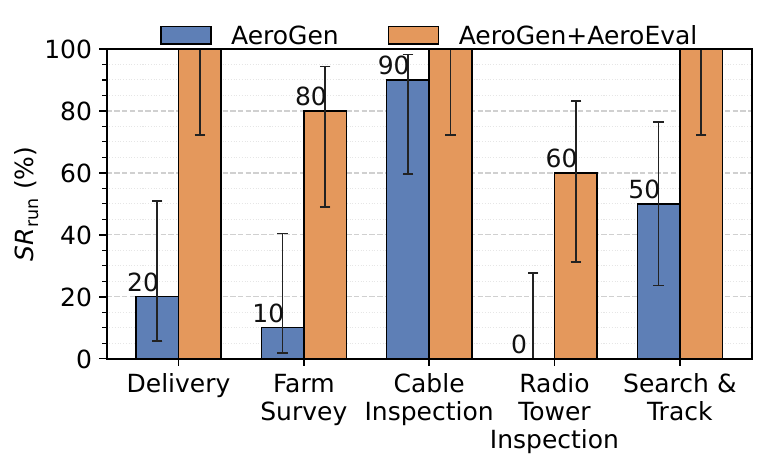}
    }
    \caption{End-to-end mission reliability using o3-mini.
    Panel~(a) compares native \CLG with \CLG+\aeval (AE) over 20 navigation
    tasks. Panel~(b) reports $SR_{\mathrm{run}}$ for five analytical
    mission types over 10 independent runs per mission; error bars show 95\% confidence intervals.}
    \label{fig:end-to-end-reliability}
\end{figure}

\subsubsection{\aeval improves convergence for navigation tasks that already
use simulator feedback.}
Native \CLG completes $11/20$ navigation tasks, for a success rate of
$55\%$, while \CLG+\aeval (AE) completes $19/20$ tasks, for a success rate
of $95\%$ (Fig.~\ref{fig:navigation-reliability}). \aeval
successfully completes the nine tasks that fail under native \CLG.
However, one task that succeeds under \CLG does not converge under
\CLG+\aeval, resulting in a net gain of eight completed tasks.

The remaining \aeval failure is Task~20, for which the generator
does not produce a program that both reaches the target altitude and
maintains the required yaw while traversing the square trajectory within
the 15-iteration budget. Successful tasks require a mean of $3.3$
iterations with \aeval compared with $2.1$ iterations for native
\CLG. Thus, staged diagnosis increases refinement effort but converts
most previously unsuccessful tasks into accepted missions.

\subsubsection{\aeval improves run-level reliability across the analytical missions.} Figure~\ref{fig:analytical-run-success} reports $SR_{\mathrm{run}}$ for 10 independent runs of each analytical mission using o3-mini. AeroGen completes $2/10$ Delivery, $1/10$ Farm Survey, $9/10$ Cable Inspection, $0/10$ Radio Tower Inspection, and $5/10$ Search and Track runs. \aeval increases these outcomes to $10/10$, $8/10$, $10/10$, $6/10$, and $10/10$, respectively. Aggregated across the five mission types, AeroGen completes $17/50$ runs ($34\%$), while \aeval completes $44/50$ runs ($88\%$), an increase of 54 percentage points.

The Wilson score $95\%$ confidence interval is $[0.22,0.48]$ for
AeroGen and $[0.76,0.94]$ for \aeval. Per-mission intervals remain
wide because each estimate uses 10 runs; therefore, we report raw counts
together with the intervals in Fig.~\ref{fig:analytical-run-success}.

Radio Tower Inspection has the lowest \aeval success because it combines search-region coverage, altitude and lane-spacing constraints, target detection, inspection, and resumption of the search pattern. In contrast, Cable Inspection has a more constrained analytical and navigation workflow. 
Across the analytical missions, the number of iterations ranges from $1$ to $15$, with Radio Tower Inspection requiring the highest mean refinement count of $12.8$ iterations across its $10$ independent runs.

\subsubsection{Candidate-level acceptance is distinct from run-level reliability.} Across the candidates generated during the \aeval refinement runs, \agen produces $13/60$ accepted Delivery, $9/75$ Farm Survey, $10/11$ Cable Inspection, $6/128$ Radio Tower Inspection, and $10/19$ Search and Track candidates. Aggregated across these mission types, $48/293$ candidates ($16.4\%$) satisfy the complete mission-level acceptance criteria.

These denominators vary because difficult missions require more refinement iterations and therefore produce more candidates. $SR_{\mathrm{candidate}}$ characterizes the individual generated programs observed during refinement, whereas $SR_{\mathrm{run}}$ measures whether an independent mission request eventually produces an accepted program within the configured budget. Candidate outcomes within the same run are not independent; therefore, we use $SR_{\mathrm{candidate}}$ as a descriptive lifecycle statistic rather than a run-level reliability estimate. \aeval accepts $44/293$ candidates, giving $SR_{\mathrm{candidate}}=15.02\%$.

\subsection{Failure Localization and Physical Applicability}
\label{subsec:localization-physical}

\begin{figure}[t]
    \centering
\subfloat[Aggregate phase-transition flow of analytical missions through the \aeval lifecycle with o3-mini]{\label{fig:sankey-diagram-agg-o3}
\includegraphics[width=0.65\columnwidth]{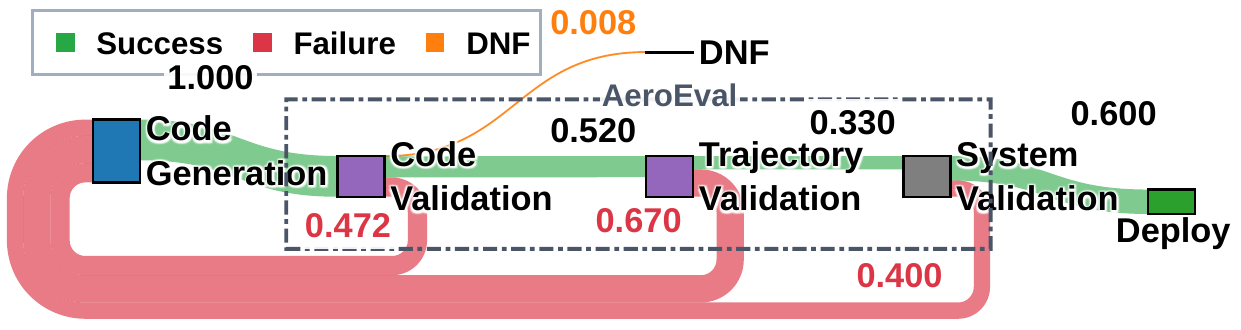}
    }\\
\subfloat[Real-world trajectories.\label{fig:traj-real-world}]{        \includegraphics[width=0.37\columnwidth]
    {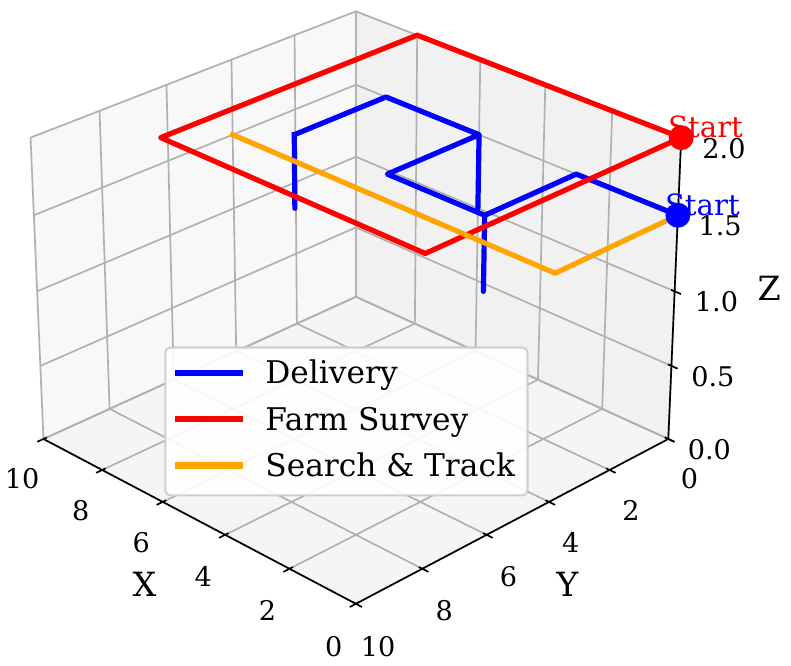}
    }\qquad
    \subfloat[Real-world farm survey]{\label{fig:real-world-survey}
\includegraphics[width=0.37\columnwidth]{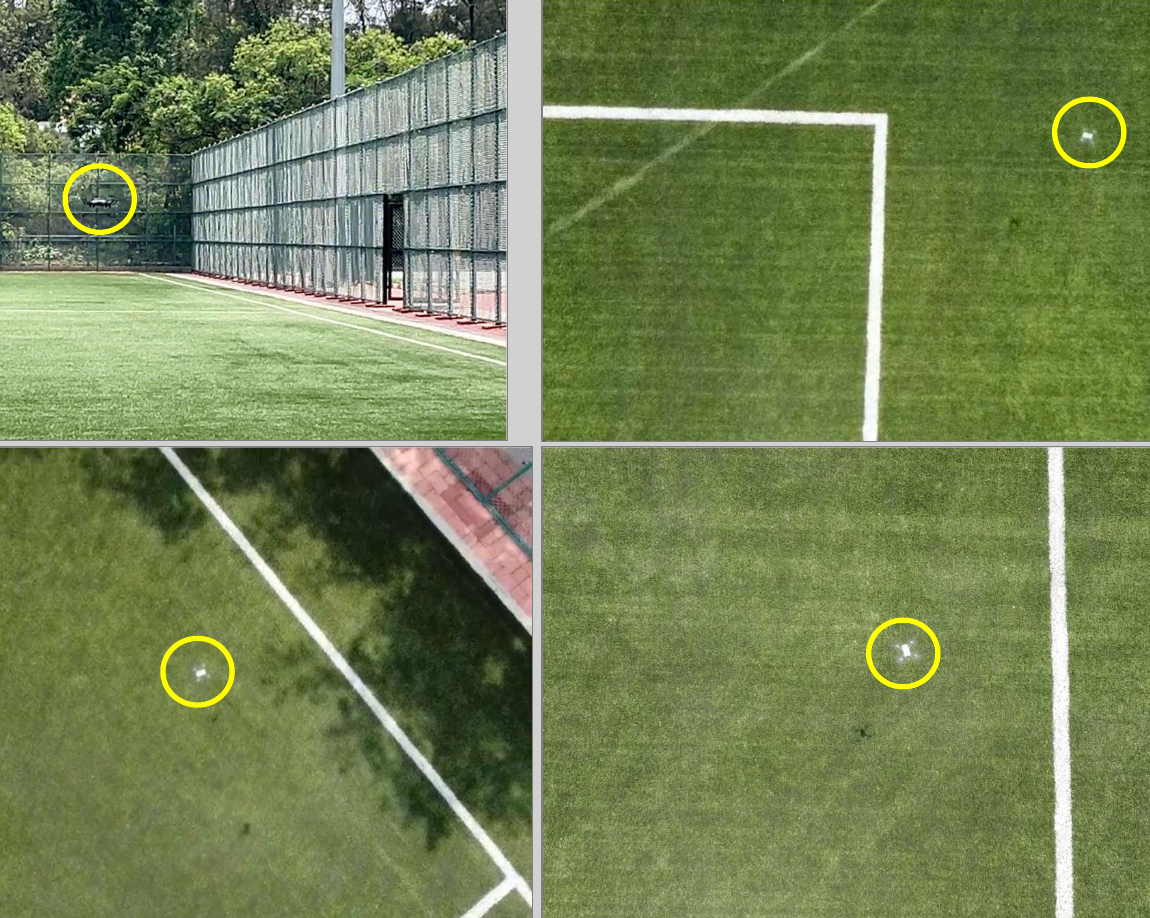}%
    }
    \caption{Failure localization and physical applicability of
    \aeval. Panel~(a) shows measured transitions of generated analytical
    candidates through the validation lifecycle using o3-mini; backward
    flows indicate regeneration and DNF denotes runs that do not
    converge within the refinement budget. Panel~(b) shows representative
    simulated and physical trajectories for missions accepted by
    \aeval and executed on a DJI Ryze Tello under expert supervision.}
    \label{fig:lifecycle-physical-evidence}
\end{figure}

\subsubsection{Program-level validity does not imply correct mission
behavior.}
Figure~\ref{fig:lifecycle-physical-evidence}(a) shows the measured
progression of analytical candidates through the \aeval lifecycle.
Across the evaluated candidates, $51.5\%$ ($151/293$) pass syntax, API,
and intent validation, while only $16.4\%$ ($48/293$) pass the complete
lifecycle. Thus, $103/293$ candidates appear valid at the program level
but subsequently fail during execution or behavioral validation.

Late-stage failures arise primarily from interactions with the physical
and mission context. These include obstacle and corridor violations,
incorrect altitude or orientation, incomplete spatial coverage, and
missing event-driven transitions. The Sankey also shows the backward
flows caused by stage-specific diagnosis and regeneration, and the
terminal flow for runs that do not converge. This measured attrition is
the \textit{illusion of correctness}: a generated program may be
syntactically valid and appear consistent with the mission, yet fail to
realize the required behavior in the evaluated environment.

\subsubsection{Accepted programs execute on the evaluated physical platform
under supervision.}
We additionally execute Virtual Drone Delivery, Farm Survey, and Search
and Track using a DJI Ryze Tello with one physical run per mission. Experiments are conducted in an open circular field of approximately 50 m radius. All three mission run succeed. The drone visits the designated waypoints for Delivery and Farm Survey,
and detects and follows the designated person wearing a high-visibility vest for Search and Track.
Each program first passes the mandatory \aeval stages in simulation and is then reviewed by a domain expert before physical execution. No manual intervention is required during execution beyond experimental setup and teardown. Owing to the Tello's limited battery life, trajectory extents and travel distances are scaled down for the physical runs, the mission logic and validation methodology remain unchanged from the simulation-accepted version.
Figure~\ref{fig:lifecycle-physical-evidence}(b--c) shows the physical
trajectories and Farm Survey execution.

The physical experiments provide evidence that accepted programs can
transfer to the evaluated drone and environments. However, the
experiments use one drone platform and a limited set of mission and
environment conditions. They therefore do not establish general
sim-to-real transfer or safety across other vehicles and operating
contexts.

\subsection{Practicality and Model Sensitivity}
\label{subsec:practicality}

\begin{table}[t]
    \centering
    \caption{Practicality and model sensitivity for analytical missions.
    Values summarize the five mission types over converged runs.}
    \label{tab:practicality}
    \small
    \setlength{\tabcolsep}{6.5pt}
    \begin{tabular}{@{}lcccc@{}}
        \toprule
        \bf Model & \bf $SR_{\mathrm{run}}$
        & \bf Validation latency
        & \bf Cached input
        & \bf Cost/iteration \\
        \midrule
        o3-mini
        & $44/50$ ($88\%$)
        & $49$--$585$\,s
        & $\approx83\%$
        & $\approx\$0.03$ \\

        Gemini-3.1-Pro
        & $47/50$ ($94\%$)
        & $7$--$1499$\,s
        & $\approx85\%$
        & $\approx\$0.07$--$0.14$ \\
        \bottomrule
    \end{tabular}
\end{table}

\subsubsection{Validation overhead is smaller than the end-to-end generation
and simulation cost for most missions.}
Excluding simulation, the median time for one analytical refinement
iteration using o3-mini ranges from $49$ to $585$\,s across the five
mission types (Table~\ref{tab:practicality}). Including simulation, the
end-to-end duration of a converged run ranges from approximately $115$\,s for Search and Track to $2.4$\,h for Radio Tower Inspection.
Across the five missions, the maximum observed latency contribution of the Code Validator and Trajectory Validator is $30\%$ and $25\%$, respectively; the remaining latency is dominated by code generation and simulated execution.

\begin{figure}
    \centering
    \includegraphics[width=0.65\linewidth]{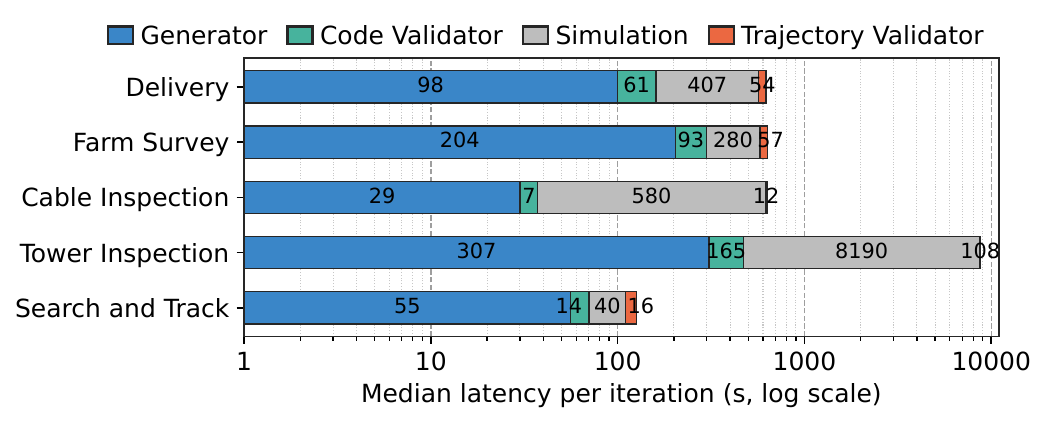}

\caption{Median end-to-end latency composition for one refinement iteration across the analytical missions using o3-mini. Labels report the total median duration, while segments show the contributions of code generation, Code Validation, simulated execution, and Trajectory Validation.}
\label{fig:latency-breakdown}
\end{figure}

Figure~\ref{fig:latency-breakdown} decomposes the median end-to-end latency of an analytical refinement iteration, excluding simulation time, into code generation, Code Validation, simulated execution, and Trajectory Validation. Simulation accounts for the largest share in long-running spatial missions, whereas code generation contributes more heavily to missions requiring complex reasoning and repair. Across the evaluated missions, the Trajectory Validator remains smaller than code generation or simulation, while the Code and Trajectory Validators contribute at most $30\%$ and $25\%$ of end-to-end latency, respectively. Thus, the additional validation stages are measurable but are not the dominant latency component for most missions.

Token use is dominated by repeated contextual input rather than newly
generated output. Across 50 runs of the five analytical missions,
o3-mini consumes $8.51$M input tokens, $0.47$M output tokens, and
$1.60$M reasoning tokens. Of the input tokens, $7.06$M are served from
the model cache, corresponding to approximately $83\%$ of the input
context. Caching therefore reduces the effective cost of repeatedly
supplying mission specifications, platform APIs, world descriptions, and
validation instructions, yielding an average cost of approximately
$\$0.03$ per analytical refinement iteration. Costs are calculated using
the standard OpenAI API prices in effect on June~6, 2026.

\subsubsection{The reliability benefit persists across the two evaluated reasoning models.} Using Gemini-3.1-Pro, \agen succeeds in $21/50$ independent runs ($42\%$), while \aeval succeeds in $47/50$ runs ($94\%$), an increase of 52 percentage points. The corresponding Wilson score $95\%$ CIs are $[0.29,0.56]$ and $[0.84,0.98]$, respectively. \aeval succeeds in $44/50$ runs ($88\%$) using o3-mini and $47/50$ runs ($94\%$) using Gemini-3.1-Pro. Since the \aeval confidence intervals overlap, the results do not establish that one model is more reliable; instead, they show that staged validation improves run-level success with both evaluated model families.
The corresponding latency, token-use, and effective-cost comparison is reported in Table~\ref{tab:practicality}.

\subsubsection{Optional deployment advice}

After mandatory validation, an optional advisor uses resource telemetry
to recommend alternative analytics placements or waypoint schedules~\cite{khochareTON}.
Its recommendations do not determine mission acceptance and are
evaluated only as an extensibility mechanism under constructed resource
contention.

\subsubsection{Summary}
The evaluation shows that program and behavioral validation detect
complementary failures, improve end-to-end mission convergence, and
localize failures that survive code-level checks. The physical
experiments provide scoped evidence of applicability on one supervised
drone platform, while the overhead results show that validation cost is
mission and model dependent. These findings apply to the evaluated
missions, models, simulators, and physical environments rather than
constituting a general assurance guarantee.

\section{Conclusion and Future Works}
\label{sec:conclusion}

We presented \aeval, an agent-assisted middleware that combines program-level and execution-grounded validation for AI-generated drone missions. \aeval separates syntax, API, and intent checks from post-simulation behavioral validation, and returns stage-specific diagnoses for iterative regeneration. For analytical missions without a reference trajectory, its Context-Grounded Agent evaluates spatio-temporal and event-driven properties using the mission specification, program, environment description, and execution observations.

Across 20 navigation tasks, \aeval increases success from $11/20$ ($55\%$) to $19/20$ tasks ($95\%$). Across five analytical mission types, it increases aggregate run-level success from $17/50$ ($34\%$) for one-shot \agen to $44/50$ ($88\%$) using o3-mini. The stagewise ablation explains this improvement: Code Validator-only and Trajectory Validator-only configurations achieve mean run-level success rates of $44\%$ and $56\%$, respectively, while their composition achieves $88\%$. This is because program validation detects structural, API, and intent failures, whereas behavioral validation detects violations involving obstacles, altitude, coverage, orientation, and mission transitions. The measured lifecycle further shows that code-valid candidates may remain behaviorally invalid. Supervised executions of three mission types on a physical drone provide limited evidence that accepted programs can transfer to the evaluated platform and environments.

\aeval provides empirical validation evidence rather than formal
guarantees of safety, liveness, or adversarial robustness. The current
evaluation is limited to two generators, two simulators, two reasoning
models, five analytical mission types, and one physical drone platform.
Moreover, using the same model for generation and validation may produce
correlated errors. Future work will examine broader mission and hardware
conditions, cross-model validation, adversarial specifications, and online monitoring that re-validates mission behavior as the execution context changes.

\section*{Acknowledgment}
Generative AI tools were solely used to refine the
clarity of author-written text in the paper. No new technical content, data or figures were generated. All such edits were carefully reviewed and verified by the authors.

\bibliographystyle{plain}
\bibliography{main}

\end{document}